\pdfoutput=1

\documentclass[11pt]{article}

\usepackage[final]{acl}

\usepackage{times}
\usepackage{latexsym}
\usepackage{booktabs}
\usepackage{graphicx} 
\usepackage{amsmath} 
\usepackage{multirow} 
\usepackage{arydshln} 

\usepackage[T1]{fontenc}

\usepackage[utf8]{inputenc}

\usepackage{microtype}

\usepackage{inconsolata}

\usepackage{graphicx}
\usepackage{amsmath}
\usepackage{amsfonts}
\usepackage{amssymb}
\usepackage{algorithm}
\usepackage{algpseudocode}
\usepackage{multirow}
\usepackage{colortbl}
\usepackage{color}
\usepackage{pifont}
\title{Autoregressive Multi-trait Essay Scoring via Reinforcement Learning with Scoring-aware Multiple Rewards}

 \author{Heejin Do$^{1}$, Sangwon Ryu$^{1}$, Gary Geunbae Lee$^{1,2}$ \\
  \centering
  \begin{tabular}[t]{c}
    $^{1}$Graduate School of Artificial Intelligence, POSTECH, South Korea \\
    $^{2}$Department of Computer Science and Engineering, POSTECH, South Korea \\
    \texttt{\{heejindo, ryusangwon, gblee\}@postech.ac.kr} \\
  \end{tabular}
}

\begin{document}
\maketitle

\begin{abstract}
Recent advances in automated essay scoring (AES) have shifted towards evaluating multiple traits to provide enriched feedback. Like typical AES systems, multi-trait AES employs the quadratic weighted kappa (QWK) to measure agreement with human raters, aligning closely with the rating schema; however, its non-differentiable nature prevents its direct use in neural network training. In this paper, we propose Scoring-aware Multi-reward Reinforcement Learning (SaMRL), which integrates actual evaluation schemes into the training process by designing QWK-based rewards with a mean-squared error penalty for multi-trait AES. Existing reinforcement learning (RL) applications in AES are limited to classification models despite associated performance degradation, as RL requires probability distributions; instead, we adopt an autoregressive score generation framework to leverage token generation probabilities for robust multi-trait score predictions. Empirical analyses demonstrate that SaMRL facilitates model training, notably enhancing scoring of previously inferior prompts.

\end{abstract}

\section{Introduction}

An essay can be evaluated from diverse perspectives, such as \textit{Content}, \textit{Sentence Fluency}, and \textit{Organization}. As providing multi-view assessment is essential for enhancing the learner's writing skill, recent attention to automated essay scoring (AES) systems has shifted from solely relying on holistic scoring \cite{taghipour2016neural, dong2016automatic, dong2017attention, wang2022use} to evaluating multiple trait scores \cite{kumar2022many, ridley2021automated, do-etal-2024-autoregressive}. Although simultaneous assessment for multiple traits is more challenging than a holistic paradigm, it has been much less explored.

\begin{figure}[t]
\centering
\includegraphics[width=0.95\linewidth]{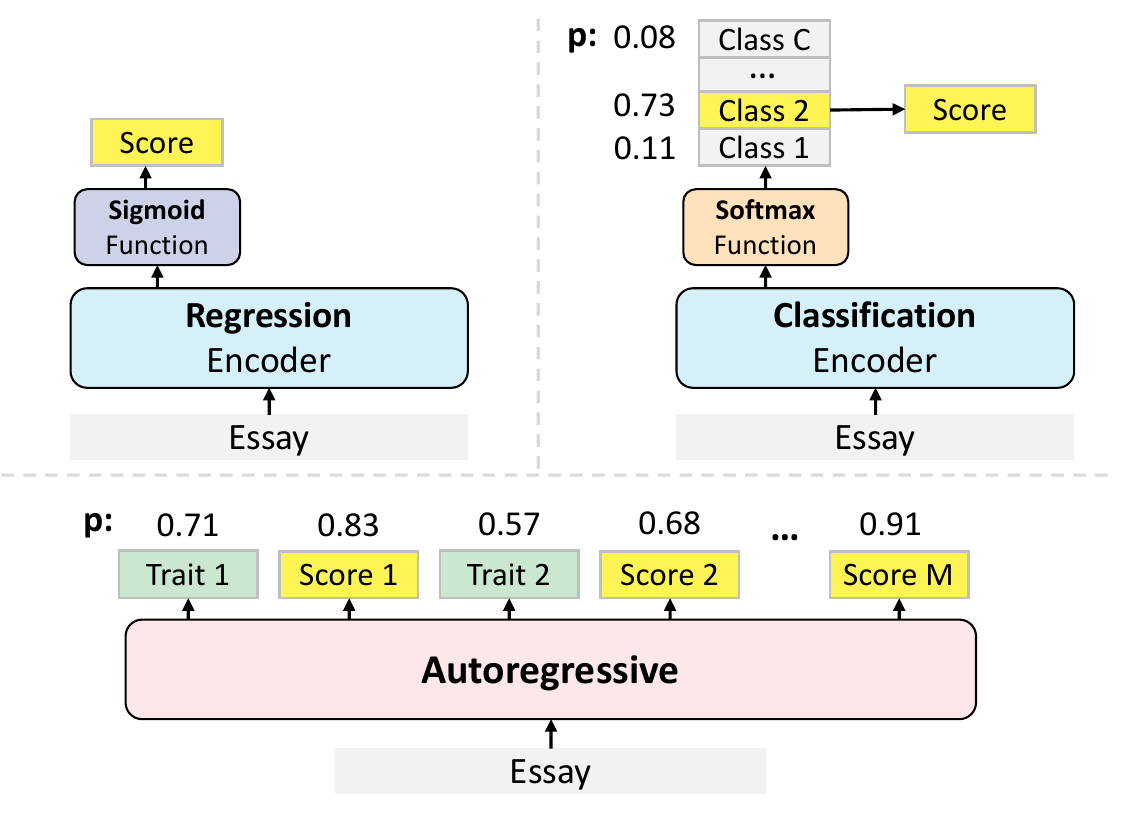}
\caption{Overview of distinct AES frameworks. The autoregressive framework eliminates the need for multiple trait-wise layers. Classification and autoregressive AES models probabilistically predict final scores; hence, a policy gradient reinforcement algorithm is applicable.}
\label{fig: front}
\end{figure}

Typically, AES systems are evaluated using the Quadratic Weighted Kappa (QWK) \cite{cohen1968weighted} score, which measures the agreement between human ratings and model predictions. Despite its effectiveness and close alignment with real-world rating schemes, its non-differentiable nature prevents direct use in neural-network training \cite{wang-etal-2018-automatic}. Instead, previous AES models predominantly utilized cross-entropy or mean squared error (MSE) loss to train classification- or regression-based AES models, respectively (Figure~\ref{fig: front}).

In this paper, we propose a Scoring-aware Multi-reward Reinforcement Learning (SaMRL) method to unlock the potential of the QWK for training multi-trait AES systems. By constructing multiple rewards of bi-directional QWK and MSE penalty, SaMRL effectively incorporates nuanced measurement schemes in training phrases. Generally, the QWK score derives from a set of essays rather than a single score; thus, when applied conventionally in a batch set, it assigns the same metric to every sample. To ensure stable training, we introduce trait-wise comparison for QWK, integrating them with the batch-wise calculation to construct a bi-directional QWK reward.

Applying RL in AES is underexplored, and prior work \cite{wang-etal-2018-automatic} is limited to holistic scoring. Further, their method is restricted to the classification approach despite its inferior performance than the regression, as policy gradients for RL require probability distributions. Unlike prior works, we treat AES as a generation paradigm \cite{do-etal-2024-autoregressive}, leveraging token generation probability distributions instead of categorical ones for policy gradient. Note that standard autoregressive AES is trained with cross-entropy and does not reflect any score-related metrics (i.e., MSE or QWK) during training; however, SaMRL enables direct parameter updates based on those scoring-aware metrics.

Extensive experiments on the representative ASAP and ASAP++ datasets demonstrate scoring enhancement over robust baselines across the traits and prompts. Comprehensive Analyses, which compare SaMRL with both single-reward applications and unidirectional use of QWK rewards, further reveal the robustness of our method. Notably, significant improvements observed on prompts with a broader score range highlight the overcoming of challenges posed by prior use of RL in AES.

\section{Related works}

\paragraph{Multi-trait essay scoring}
Although automated essay scoring has achieved notable success \cite{dong2017attention, yang2020enhancing, wang2022use}, research on multi-trait essay scoring is still underdeveloped and requires further exploration. Early attempts for multi-trait AES lie in constructing multiple trait-specific layers or models for different predictions \cite{mathias2020can, ridley2021automated, kumar2022many, do2023prompt}. Pointing out the inefficiency of duplicating individual encoder-only models generating a single score, 
\citet{do-etal-2024-autoregressive} sequentially produces multi-trait scores by defining scoring as a decoder-introduced text generation. Autoregressively generating multi-trait scores significantly improved all trait-scoring performance, achieving stat-of-the-art results. Further, it reduces the burden of designing separate trait-specific layers as it consecutively predicts full trait scores for entire prompts with a single model. To take advantage of the efficiency and high performance, we introduce our RL method based on their autoregressive AES paradigm.


\paragraph{RL for text generation}
Recently, RL has been actively employed across diverse natural language generation tasks. Notably, the advent of reinforcement learning from human feedback (RLHF) has improved the capabilities of general-purpose large language models (LLMs) such as GPT, showing the strength of RL \cite{ouyang2022training}. Numerous researchers have applied RL to specific downstream tasks such as text summarization \cite{paulus2017deep, dong-etal-2018-banditsum, chen-bansal-2018-fast, narayan-etal-2018-ranking, pasunuru-bansal-2018-multi, gunasekara-etal-2021-using-question, parnell-etal-2022-multi, roit-etal-2023-factually, ribeiro-etal-2023-generating, 9868151, ryu2024multi, singh-etal-2024-eros-entity}, machine translation \cite{wu-etal-2018-study, he2024improving}, and reasoning \cite{havrilla2024teachinglargelanguagemodels, dutta2024frugal, xi2024traininglargelanguagemodels, lu2024stepcontrolleddpoleveragingstepwise}. They aim to enhance performance by using rewards tailored to their specific objectives. For instance, in text summarization, \citet{stiennon2022learning} employ human feedback as a reward model to generate summaries that align with human preferences, while \citet{roit-etal-2023-factually} use the entailment relationship between summary and source document as a reward to generate a factually consistent summary. In arithmetic reasoning problems, \citet{dutta2024frugal} utilizes a non-differentiable symbolic solver as a reward to address multi-step reasoning. We integrate the RL framework to unexplored autoregressive multi-trait AES by introducing novel multiple scoring-aware rewards.




\section{Preliminary}
We adopt the policy gradient reinforcement learning to train the policy to optimize rewards. Policy gradient aims to increase the probability of actions that yield high rewards. To guide the model towards taking actions that result in a higher reward, the policy gradient function includes the probability of actions taken by the policy, $\pi_\theta (a \mid s)$:
\begin{equation} \label{eq: pgloss}
L^{PG}(\theta) = \hat{E}_t \left[ \log \pi_\theta (a_t \mid s_t) \hat{A}_t \right]
\end{equation}

Most of the existing AES systems use encoder-only models like BERT \cite{devlin-etal-2019-bert} and rely on regression- or classification-based approaches. In general, the model generates the essay embedding vector from a given essay input; however, the differences lie in the objective functions and the process of deriving the final score from the output vector. Regression models predict the score with a sigmoid function given the embedding vector and are trained with the MSE loss function between the label $y$ and predicted score $\hat{y}$ for $n$ samples:




\begin{equation} \label{eq: mseloss}
loss_{MSE} = \frac{1}{n} \sum_{i=1}^{n} (y - \hat{y})^2
\end{equation}




\begin{figure*}[t]
\centering
\includegraphics[width=0.99\textwidth]{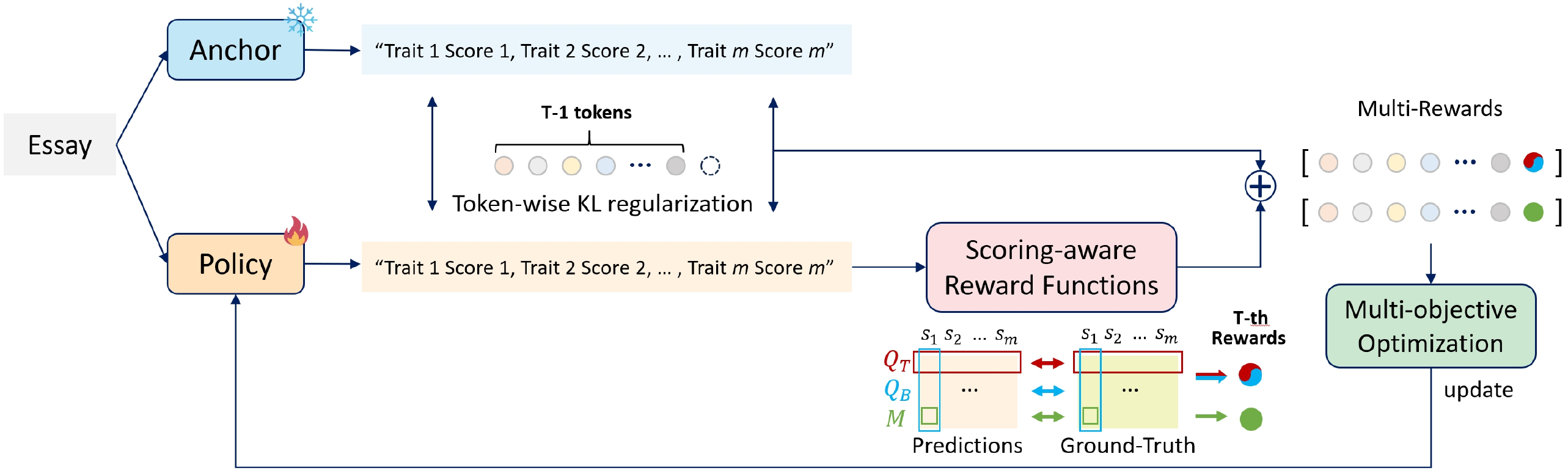}
\caption{Overview of the entire process for the proposed autoregressive multi-trait AES with SaMRL. We maintain the structure of the score generation within the policy model through token-wise KL regularization and allow the model to align with human judgment by introducing multiple scoring-aware rewards.}
\label{fig: main}
\end{figure*}

However, as they output a single scalar value, they are unsuitable for policy gradient, which requires a probability distribution for policy training. Contrarily, classification-based AES models output a probability distribution ($p$) given the essay vector with the softmax function, and cross-entropy loss is employed as training objectives:
\begin{equation}
loss_{CE} = -\sum_{i=1}^{C} y_i \log(p_i)    
\end{equation}
where $C$ denotes the number of classes, and $y_i$ is a one-hot vector having one for the ground-truth category. By treating scoring a single essay as an action, the RL mechanism can be applicable to classification models \cite{wang-etal-2018-automatic}. However, classification-based AES typically underperforms regression approaches, particularly lagging behind in prompts with broader score ranges \cite{wang-etal-2018-automatic}. This performance drop is attributed to the increased number of prediction candidate classes. 


To eliminate a performance decline, we introduce the RL application for AES with a generation framework. In particular, we ground our method on an autoregressive multi-trait score generation model, ArTS \cite{do-etal-2024-autoregressive}, instead of classification approaches. We consider generating text trait and score tokens as an action and treat token generation probability distribution as the probability of actions. In addition, to take advantage of comparing the exact error rate in regression, we leverage the MSE penalty as part of our multi-rewards. 







\section{SaMRL}
To incorporate a rating schema in the training phase, SaMRL updates the policy model using multiple rewards obtained by a scoring-aware multi-reward function. In RL, relying solely on reward-based learning can lead the model to prioritize enhancing rewards exclusively, potentially losing its capacity to score essays in appropriate prediction form. Therefore, we employ a fixed-parameter anchor model to guide the policy to prevent significant deviations in training and maintain the trait patterns. Figure~\ref{fig: main} describes the overall process of SaMRL. 

\subsection{Score generation model}

In autoregressive multi-trait prediction, given the essay input with the prefix of \textit{"score the essay of the prompt $i$: "}, the model outputs the sequence of the trait name and the corresponding score, such as $[trait_1 \ score_1, trait_2 \ score_2, …, trait_m \ score_m]$. In this work, we denote the generated sequence as $\hat{y}$ and the ground-truth sequence as ${y}$; the extracted numeric trait scores from each are denoted as $\hat{s}$ and $s$, respectively. The sequence-to-sequence T5 \cite{2020t5} model is employed for score generation, adhering to the same trait prediction order as the baseline model \cite{do-etal-2024-autoregressive}. This order progresses from traits with smaller data sizes to those with larger data sizes. Also, traits that are not evaluated in certain prompts are regarded and predicted as the \textit{"nan"} value, such as $[trait_j \ nan]$. 

\subsection{Multi-rewards function} 
To manage both the overall consistency and agreement across ratings, as well as precision at individual score levels, we introduce multiple rewards: bidirectional QWK ($r_{Q}$) and mean trait-wise MSE reward ($r_{M}$). QWK accounts for the ordinal nature of essay scores and weighting exact and near matches; thus, it effectively captures the rating schema and is sensitive to the overall qualities \cite{wang-etal-2018-automatic}. Meanwhile, MSE aggregates the exact difference between predicted and actual scores; hence, it provides a clear yet simple indication of individual score deviations from true labels.  


\paragraph{QWK}
The QWK metric evaluates the agreement and consistency between human and system assessments across a set of essays, calculating a single QWK score for the entire set. Therefore, rewarding with the measurement within the batch set, $Q_{B}$, assigns the same reward to all samples, potentially resulting in unstable training. To establish a more precise and stable reward strategy, we introduce a trait-wise QWK, $Q_{T}$, which evaluates the agreement between the trait sets of predictions and gold labels at the sample level. Specifically, batch-wise quadratic weighted kappa score, $Q_{B}$, is defined as: 
\begin{eqnarray}
    Q_{B} = 1 - \frac{\sum_{i,j} W_{i,j}C_{i,j}}{\sum_{i,j} W_{i,j}E_{i,j}}
\end{eqnarray}
where the $N \times N$ weight matrix $W_{i,j}$ is calculated as ${(i-j)^2}/{(N-1)^2}$ for the number of candidate ratings $N$. $C_{i,j}$ denotes the counts of the essays assigned $i$ score by human grader, and $j$ score by our system. $E_{i,j}$ represents the expected count of two ratings assigning the $i$ and $j$ scores, respectively, and calculated by the outer product of two ratings' histogram vectors \cite{wang-etal-2018-automatic}. Akin to the batch-wise measurement, the sample-level trait-wise QWK, $Q_{T}$, is computed using three metrics, but the focus is on the set of traits within a single sample rather than the essay set within a batch. Particularly, $C_{i,j}$ denotes the number of traits received a score of $i$ by a human grader and a score of $j$ by the system. Then, the bidirectional QWK reward $r_Q$, integrating two-way calculation, is defined as:
\begin{equation}
    r_Q(S,\hat{S}) = \lambda \cdot Q_B + (1-\lambda)\cdot Q_T
\end{equation}\label{lambda}
for in-batch prediction and actual score vectors $\hat S$ and $S$, respectively. The prior RL for AES used the packed evaluation \cite{wang-etal-2018-automatic}, averaging the QWK obtained from randomly selected essay packs. Conversely, our bidirectional approach eliminates the need for multiple assemblies per every target essay and removes random variability, achieving both efficient and reliable rewards.

\begin{table*}[ht]
\centering
\scalebox{
0.73}{
\begin{tabular}{cccccccc}
\toprule
\multirow{2}{*}{\textbf{Prompt}} & \multirow{2}{*}{\textbf{\# of Essays}} & \multirow{2}{*}{\textbf{Average Length}} & \multirow{2}{*}{\textbf{Essay Type}} & \multirow{2}{*}{\textbf{Grade Level}} & \multirow{2}{*}{\textbf{Traits}} & \multicolumn{2}{c}{\textbf{Score Range}} \\
 &  &  &  &  &  & \textbf{Overall} & \textbf{Trait} \\
\midrule
P1 & 1,783 & 350 & Argumentative & 8 & Over, Cont, Org, WC, SF, Conv & 2 - 12 & 1 - 6 \\
P2 & 1,800 & 350 & Argumentative & 10 & Over, Cont, Org, WC, SF, Conv & 1 - 6 & 1 - 6 \\
P3 & 1,726 & 150 & Source-Dependent & 10 & Over, Cont, PA, Lan, Nar & 0 - 3 & 0 - 3 \\
P4 & 1,772 & 150 & Source-Dependent & 10 & Over, Cont, PA, Lan, Nar & 0 - 3 & 0 - 3 \\
P5 & 1,805 & 150 & Source-Dependent & 8 & Over, Cont, PA, Lan, Nar & 0 - 4 & 0 - 4 \\
P6 & 1,800 & 150 & Source-Dependent & 10 & Over, Cont, PA, Lan, Nar & 0 - 4 & 0 - 4 \\
P7 & 1,569 & 300 & Narrative & 7 & Over, Cont, Org, Conv, Style  & 0 - 30 & 0 - 6 \\
P8 & 723 & 650 & Narrative & 10 & Over, Cont, Org, WC, SF, Conv, Voice  &  0 - 60 & 2 - 12 \\
\bottomrule
\end{tabular}}
\caption{ASAP and ASAP++ combined dataset statistics. Over: \textit{Overall}, Cont: \textit{Content}, Org: \textit{Organization}, WC: \textit{Word Choice}, SF: \textit{Sentence Fluency}, Conv: \textit{Conventions}, PA: \textit{Prompt Adherence}, Lang: \textit{Language}, Nar: \textit{Narrativity}.}
\label{tab: dataset statistics}
\end{table*}

\paragraph{MSE}
The autoregressive generation framework for scoring, which bases its predictions on token probabilities, does not involve direct numerical comparisons between predicted scores and human labels, such as error rates between ratings. We introduce MSE rewards as an auxiliary component to allow a generation-based scoring system to incorporate quantitative comparisons for ratings. As we consider multiple trait scores, we use mean trait-wise root mean squared error, which computes the average of the squared errors for each trait. The MSE reward, $r_M$ is defined as follows:
\begin{equation}
    r_M(S,\hat{S}) = - \frac{1}{m} \sum_{j=1}^{m} \sqrt{\frac{1}{n} \sum_{i=1}^{n} (s_{ij} - \hat{s}_{ij})^2}
\end{equation}
for $m$ number of traits and $n$ number of predicted samples. 


\subsection{RL policy update}
Our multi-trait AES model generates scores sequentially in a specific order and format. Consequently, a comprehensive evaluation is only feasible after the complete sequence has been generated, culminating with the final trait $[trait_m\  score_m]$. Thus, the model receives calculated multi-rewards just after the completion of the last trait score generation at the $T$-th timestep. Maintaining a structured format is crucial for scoring multiple traits in order, we adopt the token-wise KL regularization technique between the frozen anchor model, $\pi^{\mathrm{AC}}$, and our trainable policy model $\pi^{\mathrm{\theta}}$. This process prevents the policy from overly adapting to the reward function while ensuring the preservation of the $[trait_j\ score_j]$ generation format. For each multi-rewards, in conjunction with the obtained reward and token-wise KL regularization until the $T-1$-th tokens, we update the policy via PPO \cite{schulman2017proximal} with generalized advantage estimation \cite{schulman2018highdimensional}. The full reward $R_k$ ($k \in \{M,Q\}$), is defined as follows:
\begin{equation}
\mathrm{R_k}(e, y, \hat{y}) = r_k(y, \hat{y}) - \beta \log\left[\frac{\pi_{\theta}^{\mathrm{RL}}(\hat{y_t}|e)}{\pi^{\mathrm{AC}}(\hat{y_t}|e)}\right]
\end{equation}

where $e$ is the essay, $\hat{y_t}$ is the $t$-th token in the model-generated sequence $\hat{y}$, and $y$ is the reference sequence. The policy $\pi_\theta$ is the autoregressive multi-trait AES model, and $e$ is the essay. In our work, the action denotes the selection of the next token by the policy. The action space $V$ corresponds to the vocabulary of the policy $\pi$. Then, the PPO loss using $R_k$ is defined as, 
\begin{equation}
\scriptsize loss^{k}_{t}(\theta) = \hat{\mathbb{E}}_t \Bigl[loss^{\mathrm{CLIP}}_t(\theta) - c_1 loss^{\mathrm{VF}}_t (\theta) + c_2 H[\pi_\theta](y_t) \Bigr]
\end{equation}
where $loss^{\mathrm{CLIP}}$ represents the constrained surrogate loss through clipping, while the $loss^{\mathrm{VF}}$ denotes the mean squared error update of the value function. $H$ stands for entropy. 








\paragraph{Multi-objective Optimization} 
We consider two distinct multi-rewards--$R_Q$ and $R_M$--for individually computing the PPO loss. Considering each loss as $loss_{R_{Q}}$ and $loss_{R_{M}}$, we treat the training process as multi-task learning with this formula:
\begin{equation}
    loss_{total} = w_{Q}loss_{R_{Q}} + w_{M}loss_{R_{M}}
\end{equation}

To dynamically optimize multiple loss functions along with their respective weights, we also train and update $w_Q$ and $ w_M$ rather than relying on static interpolation. They are normalized with the softmax function. 


\section{Experimental setup}
\paragraph{Datasets}
We use the open-sourced ASAP\footnote{https://www.kaggle.com/c/asap-aes} and ASAP++\footnote{https://lwsam.github.io/ASAP++/lrec2018.html} \citep{mathias2018asap++} datasets, as in the baseline ArTS model \cite{do-etal-2024-autoregressive}. ASAP++ dataset includes enriched human-annotated multi-trait scores for English-written essays of eight separate prompts. As summarized in Table~\ref{tab: dataset statistics}, different prompts are assessed with distinct traits with varied score ranges. While other traits are evaluated across several prompts, \textit{Style} and \textit{Voice} are only assessed in prompts 7 and 8, respectively, resulting in a very limited number of samples available for training. Our model handles all prompts and all traits with a single model. We also experiment with the publicly available Feedback Prize \footnote{https://www.kaggle.com/competitions/feedback-prize-english-language-learning} dataset, where argumentative essays are annotated with six trait scores, \textit{Cohesion}, \textit{Syntax}, \textit{Vocabulary}, \textit{Phraseology}, \textit{Grammar}, and \textit{Conventions}. Unlike the ASAP dataset, this data is not divided by prompts.

\paragraph{Evaluations}
Following the previous works \cite{taghipour2016neural, do-etal-2024-autoregressive}, we use five-fold cross-validation using the same split as their works. The ASAP dataset's official metric, QWK, is used as the evaluation metric, and five-fold averaged values and standard deviations are reported. We also measure separate QWK by prompts to ensure fair comparisons and prevent overly high scores in case of testing in the entire set \citet{taghipour2016neural, do-etal-2024-autoregressive}. Comparison models are detailed in Appendix~\ref{Comparison models}.

\paragraph{Settings}
For the experiments, we use the T5-Base and T5-Large \cite{2020t5} models as the policy and anchor models, which are based on the Transformer architecture. We set the generation hyperparameters the same as the ArTS, using Seq2SeqTrainer with 5000 evaluation steps, 2 early stopping patients, 4 batch sizes, and 15 total epochs. For RL hyperparameters, we mainly follow previous RL-based models \cite{dutta2024frugal, roit-etal-2023-factually, ryu2024multi}, adopting batch size of 4, discount factor $\gamma$ as 0.99 and a learning rate of 1.41e-6. For $r_Q$ calculation in Equation~\ref{lambda}, we use 0.5 as $\lambda$. \texttt{A100-SMX4-8} and \texttt{A6000} GPUs are used. 


\section{Result and Discussions}

\begin{table*}[t]
\centering
\scalebox{
0.73}{
\begin{tabular}{l|ccccccccccc|c}
\toprule
& \multicolumn{11}{c|}{\textbf{Traits (Prediction Order: ←)}} & \\
\hline
\textbf{Model} &  Overall & Content & PA & Lang & Nar & Org & Conv & WC & SF & Style & Voice & AVG↑ (SD↓)\\
\hline
HISK & 0.718 & 0.679 & 0.697 & 0.605 & 0.659 & 0.610 & 0.527 & 0.579 & 0.553 & 0.609 & 0.489 & 0.611 (-) \\
STL{\small-LSTM} & 0.750 & 0.707 & 0.731 & 0.640 & 0.699 & 0.649 & 0.605 & 0.621 & 0.612 & 0.659 & 0.544 & 0.656 (-) \\
MTL{\small-BiLSTM} & {0.764} & 0.685 & 0.701 & 0.604 & 0.668 & 0.615 & 0.560 & 0.615 & 0.598 & 0.632 & {0.582} & 0.638 (-) \\
{ArTS} & 0.754 & {0.730} &{0.751} & {0.698} & {0.725} & {0.672} & {0.668} & {0.679} & {0.678} & {0.721} & 0.570 & {0.695} (±0.018) \\
\midrule
{ArTS-base*} & 0.737 & 0.727 & 0.751 & 0.702 & {0.739} & 0.665 & 0.679 & 0.672 & 0.679 & \textbf{0.735} & {0.557} & 0.695 (±0.022) \\
\textbf{SaMRL-base} (Ours) & \textbf{0.750} & \textbf{0.732} & \textbf{0.754} & \textbf{0.704} & \textbf{0.740} & \textbf{0.670} & \textbf{0.684} & \textbf{0.681} & \textbf{0.685} & 0.726 & \textbf{0.558} & \textbf{0.699} (±0.022) \\
\midrule
{ArTS-large*} & 0.752 & 0.729 & 0.749 & 0.701 & 0.727 & 0.677 & 0.683 & 0.683 & 0.683 & \textbf{0.712} & 0.621 & 0.702 (±0.013) \\
\textbf{SaMRL-large} (Ours) & \textbf{0.754} & \textbf{0.735} & \textbf{0.751} & \textbf{0.703} & \textbf{0.728} & \textbf{0.682} & \textbf{0.685} & \textbf{0.688} & \textbf{0.691} & 0.710 & \textbf{0.627} & \textbf{0.705} (±0.013)\\
\bottomrule
\end{tabular}}
\caption{\label{tab: main}
Evaluated QWK results averaged across the prompts for each \textbf{trait}. Traits are predicted from right to left (←). Five-fold averaged standard deviation is reported (\textit{SD}). ArTS* is our implemented version, and ArTS is the reported ones in \citet{do-etal-2024-autoregressive}. Higher values among the implemented baseline and ours are represented in \textbf{bold}.}

\smallskip
\smallskip

\scalebox{
0.73}{
\begin{tabular}{l|cccccccc|c}
\toprule
& \multicolumn{8}{c|}{\textbf{Prompts}} & \\
\hline
\textbf{Model} & 1 & 2 & 3 & 4 & 5 & 6 & 7 & 8 & AVG↑ (SD↓) \\
\hline
HISK & 0.674 & 0.586 & 0.651 & 0.681 & 0.693 & 0.709 & 0.641 & 0.516 & 0.644 (-) \\
STL{\small-LSTM} & 0.690 & 0.622 & 0.663 & 0.729 & 0.719 & 0.753 & 0.704 & 0.592 & 0.684 (-) \\
MTL{\small-BiLSTM}  & 0.670 & 0.611 & 0.647 & 0.708 & 0.704 & 0.712 & 0.684 & 0.581 & 0.665 (-) \\
{ArTS} & {0.708} & {0.706} & {0.704} & {0.767} & {0.723} & {0.776} & {0.749} & {0.603}& {0.717} (±0.025) \\
\midrule
{ArTS-base*} & 0.712 & 0.680 & 0.713 & 0.771 & \textbf{0.730} & 0.775 & \textbf{0.747} & 0.595 & 0.715 (±0.016) \\
\textbf{SaMRL-base} (Ours) & \textbf{0.717} & \textbf{0.703} & \textbf{0.715} & \textbf{0.773} & 0.729 & \textbf{0.778} & 0.745 & \textbf{0.604} & \textbf{0.720} (±0.015) \\
\midrule
{ArTS-large*} & 0.700 & 0.699 & 0.704 & 0.766 & \textbf{0.725} & 0.770 & 0.739 & 0.644 & 0.718 (±0.012) \\
\textbf{SaMRL-large} (Ours) & \textbf{0.702} & \textbf{0.711} & \textbf{0.708} & \textbf{0.766} & 0.722 & \textbf{0.773} & \textbf{0.743} & \textbf{0.649} & \textbf{0.722} (±0.012) \\
\bottomrule
\end{tabular}}
\caption{\label{tab: prompt}
Evaluated QWK results averaged across the traits for each \textbf{prompt}.}
\end{table*}

\paragraph{Main results}
The main experimental results for each trait in Table~\ref{tab: main} demonstrate the effectiveness of our SaMRL method in enhancing the scoring quality across most traits, establishing new state-of-the-art.
Noticeably, applying SaMRL to both ArTS-base and ArTS-large models exhibits a consistent trend of performance enhancements across the traits. We conducted a paired t-test on trait-wise average scores, and both the SaMRL-base and -large models showed significant improvements over the baseline (p < 0.05). Given that the T5-base and T5-large models have 220M and 770M parameters, respectively, these findings underscore the robustness and consistency of our method regardless of model size. Unlike the \textit{Content} or \textit{Overall} traits, which are evaluated on every prompt's essay sets and are benefited by our approach, the \textit{Style} trait did not benefit from our method. SaMRL incorporates QWK scores calculated between batches as rewards during the reinforcement learning phase; thus, traits that are more prevalent within a batch are likely to be more dominantly reflected. As \textit{Style} is evaluated only in prompt 7, and the number of corresponding essays is limited to 1,569 (Table~\ref{tab: dataset statistics}), its contribution within a batch might be overshadowed by others. 

\begin{figure}[t]
\centering
\includegraphics[width=\linewidth]{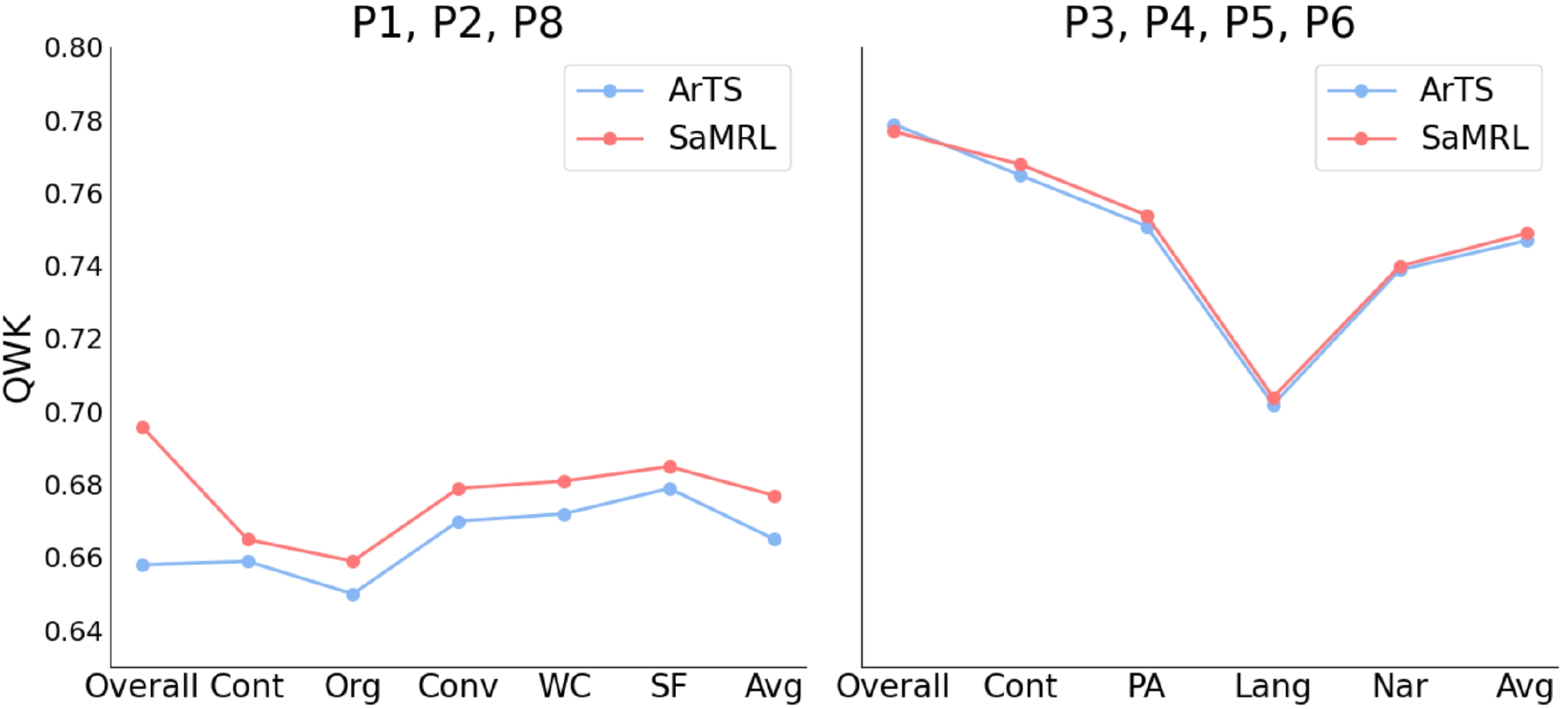}
\caption{Comparison of performance between different prompt types with varying trait compositions. Prompts 1,2 and 8 are evaluated on the same traits, while 3-6 prompts are assessed on the other same traits.}
\label{fig: prompt}
\end{figure}


Similar observations are evident in the prompt-wise results presented in Table~\ref{tab: prompt}, which displays improvements across most prompts. Like the trait results, our method consistently exhibited increasing trends across different prompts, irrespective of model size. Notably, remarkable improvements are observed in prompts 1, 2, and 8, which share the same trait sets, \textit{Cont}, \textit{Org}, \textit{WC}, \textit{SF}, \textit{Conv} as detailed in Table~\ref{tab: dataset statistics}. Consequently, we further analyze those prompts by examining the performance variations for each associated trait.



\begin{table*}[t]
\centering
\scalebox{
0.73}{
\begin{tabular}{l|ccccccccccc|c}
\toprule
& \multicolumn{11}{c|}{\textbf{Traits (Prediction Order: ←)}} & \\
\hline
\textbf{Model} &  Overall & Content & PA & Lang & Nar & Org & Conv & WC & SF & Style & Voice & AVG↑ (SD↓)\\
\hline
{ArTS-base*} & 0.737 & 0.727 & 0.751 & 0.702 & {0.739} & 0.665 & 0.679 & 0.672 & 0.679 & 0.735 & 0.557 & 0.695 (±0.018) \\
\midrule
{SaSRL$_M$} & 0.749 & 0.728 & 0.751 &	0.702 &	0.738 &	0.665 &	0.678 &	0.674 &	0.679 &	0.733 &	0.556 &  0.696 (±0.020) \\
{SaSRL$_Q$} & 0.730 & 0.720 & 0.745 & \textbf{0.704} & 0.733 & 0.660 & 0.670 & 0.664 & 0.669 & 0.733 & \textbf{0.570}  & 0.691 (±0.020) \\
{SaMRL\_uniQ$_T$} & 0.737 &	\textbf{0.733} &	0.753 &	0.701 &	{0.739} &	\textbf{0.671} &	0.679 &	0.678 &	0.683 &	\textbf{0.737} &	0.562 & {0.698} (±0.023) \\
{SaMRL\_uniQ$_B$} & 0.734 & 0.731 & 0.753 & 0.701 &	0.737 &	0.667 &	0.681 &	0.674 &	0.680 &	0.729 &	0.556 & 0.695 (±0.022) \\
\midrule
\textbf{SaMRL\_biQ} (Ours) & \textbf{0.750} & {0.732} & \textbf{0.754} & \textbf{0.704} & \textbf{0.740} & {0.670} & \textbf{0.684} & \textbf{0.681} & \textbf{0.685} & 0.726 & {0.558} & \textbf{0.699} (±0.022) \\

\bottomrule
\end{tabular}}
\caption{\label{tab: ablation_trait}
Ablation results comparing the use of score-ware single rewards (SaSRL$_M$ and SaSRL$_Q$) and the implementation of unidirectional QWK rewards (SaMRL$\_{uniQ_T}$ and SaMRL$\_{uniQ_B}$) instead of bidirectional ones. SaMRL$\_{biQ}$ denotes our SaMRL model, using multi-rewards with bidirectional QWK reward. 
}


\smallskip
\smallskip

\scalebox{
0.73}{
\begin{tabular}{l|cccccccc|c}
\toprule
& \multicolumn{8}{c|}{\textbf{Prompts}} & \\
\hline
\textbf{Model} & 1 & 2 & 3 & 4 & 5 & 6 & 7 & 8 & AVG↑ (SD↓) \\
\hline
{ArTS-base*} & 0.712 & 0.680 & 0.713 & 0.771 & 0.730 & 0.775 & 0.747 & 0.595 & 0.715 (±0.025) \\
\midrule
{SaSRL$_M$} & 0.708 & 0.699 & 0.714 & 0.770 & 0.729 & 0.774 & 0.747 & 0.599 & 0.718 (±0.015)\\
{SaSRL$_Q$} & 0.705 & 0.677 & 0.712 & 0.765 & 0.725 & 0.767 & 0.744 & 0.586 & 0.710 (±0.016)\\
{SaMRL\_uniQ$_T$} & \textbf{0.718} & 0.683 & 0.710 & \textbf{0.773} & \textbf{0.732} & 0.776 & \textbf{0.750} & 0.600 & 0.718 (±0.016)\\
{SaMRL\_uniQ$_B$} & 0.711 & 0.684 & 0.711 & 0.770 & 0.731 & 0.774 & 0.746 & 0.598 & 0.716 (±0.016)\\
\midrule
\textbf{SaMRL\_Q} (Ours) & {0.717} & \textbf{0.703} & \textbf{0.715} & \textbf{0.773} & 0.729 & \textbf{0.778} & 0.745 & \textbf{0.604} & \textbf{0.720} (±0.015) \\
\bottomrule
\end{tabular}}
\caption{\label{tab: ablation_prompt}
Evaluated QWK results averaged across the traits for each \textbf{prompt}.}
\end{table*}

\begin{table}[t]
\centering
\scalebox{
0.69}{
\begin{tabular}{c|c|c|c|c|c|c|c}
\toprule
& \multicolumn{6}{c|}{\textbf{Traits (Prediction Order: ← )}} & \\
\hline
\textbf{Model}  & Coh & Syn & Voc & Phr & Gram & Conv & AVG \\
\hline
MTL* &  0.462 & 0.507 & 0.519 & 0.505 & 0.484 & 0.527 & 0.501 \\
{ArTS}* & {0.590} & {0.628} & {0.594} & {0.639} & \textbf{0.659} & \textbf{0.659} & {0.628} \\
\midrule
{SaMRL}* & \textbf{0.592} & \textbf{0.637} & \textbf{0.610} & \textbf{0.646} & {0.658} & {0.656} & \textbf{0.633} \\
\bottomrule
\end{tabular}}
\caption{\label{tab: feedback}
Results on the Feedback Prize dataset. Five-fold averaged QWK score is reported; Coh: \textit{Cohesion}, Syn: \textit{Syntax}, Voc: \textit{Vocabulary}, Phr: \textit{Phraseology}, Gram: \textit{Grammar}, Conv: \textit{Conventions}.}
\end{table}

\paragraph{Impacts on different essay types}
Specifically, we separately investigate the trait-specific performance improvements among prompts sharing common traits, comparing prompts 1,2 and 8 with prompts 3--6. Intriguingly, Figure~\ref{fig: prompt} reveals that our model's impact varies with the evaluated prompt type. Our model, SaMRL, demonstrates more substantial performance gains for the \textit{Argumentative} or \textit{Narrative} types, such as prompts 1, 2, and 8, compared to the \textit{Source-dependent} prompts 3--6, which necessitate essay writings based on source texts. The marked difference in scoring range--where the latter prompts are evaluated on a [0, 4] scale and the former on a broader range up to [0, 60]--highlights that our model effectively addresses the challenges posed by traditional classification-based RL models, which typically struggle in wider score ranges. These advances might be attributed to direct updates of SaMRL grounded on score-difference-reflected rewards measured by QWK and MSE, which enables the inclusion of scoring schemas (e.g., imposing significant penalties for inaccurately predicting wide-range scores) that typical generation processes fail to accommodate. Moreover, the results suggest that our approach can be particularly advantageous in less favorable conditions, given that prompts 1, 2, and 8 collectively have 4.3K essay samples, while 3--6 prompts comprise 7.1K samples. 



\paragraph{Feedback Prize dataset}
To validate the broader applicability of SaMRL, we conducted additional experiments using the Feedback Prize dataset. As shown in Table~\ref{tab: feedback}, there is an overall enhancement in performance across the traits. Notably, although this dataset comprises only 2.3K training samples, substantially fewer than the over threefold larger ASAP dataset, the observed improvements support our findings of benefits in less favorable conditions.

\subsection{Ablation studies}




\paragraph{Are multi-rewards more effective than single rewards?}
We conduct comprehensive ablation studies to analyze the respective and joint effects of each reward. Experimental results in Table~\ref{tab: ablation_trait} and~\ref{tab: ablation_prompt} reveal that our multi-rewards are more effective than separately applying single rewards. When adopted individually, MSE-only reward (SaSRL$_M$) achieves higher performance than bidirectional QWK-only reward (SaSRL$_Q$). The result validates the efficacy of adjusting MSE objectives from logistic regression models to fit autoregressive frameworks via RL. Meanwhile, the joint use of both rewards (SaMRL) shows superior improvements across all prompts and traits except \textit{Style}, indicating synergistic impacts of our multi-rewarding mechanism. Our findings align with the prior works \cite{pmlr-v202-dann23a}, suggesting the effectiveness of multi-reward RL over using a single reward. 



\paragraph{Is bidirectional QWK more effective than unidirectional QWK?}
In addition, we analyze whether our bidirectional QWK strategy is indeed more advantageous than the unidirectional QWK rewards for policy training. Note that SaMRL$\_uniQ_T$ and SAMRL$\_uniQ_B$ are the models in which $Q_T$ and $Q_B$ are applied in combination with MSE rewards, respectively. We observe overall higher performances when solely using the trait-wise QWK $Q_T$ as $R_Q$ (SaMRL$\_uniQ_T$) than when only applying the batch-wise QWK, $Q_B$ (SaMRL$\_uniQ_B$). As we hypothesized, using $Q_B$ alone in a traditional essay-set-wise comparison can lead to unstable learning outcomes as it assigns the same reward to all in-batch samples. This phenomenon is also underscored by its degraded performance when compared to SaSRL$_M$, which exclusively uses MSE. Remarkably, introducing trait-set-wise unidirectional measurements $Q_T$ alone brings in significant improvements over the baseline model and SaSRL$_M$, showing even comparable results to SaMRL$\_biQ$. Nevertheless, the combined use of both directional calculations for the QWK reward stands as the most effective, implying their potential to complement each other.


\begin{figure}[t]
\centering
\includegraphics[width=\linewidth]{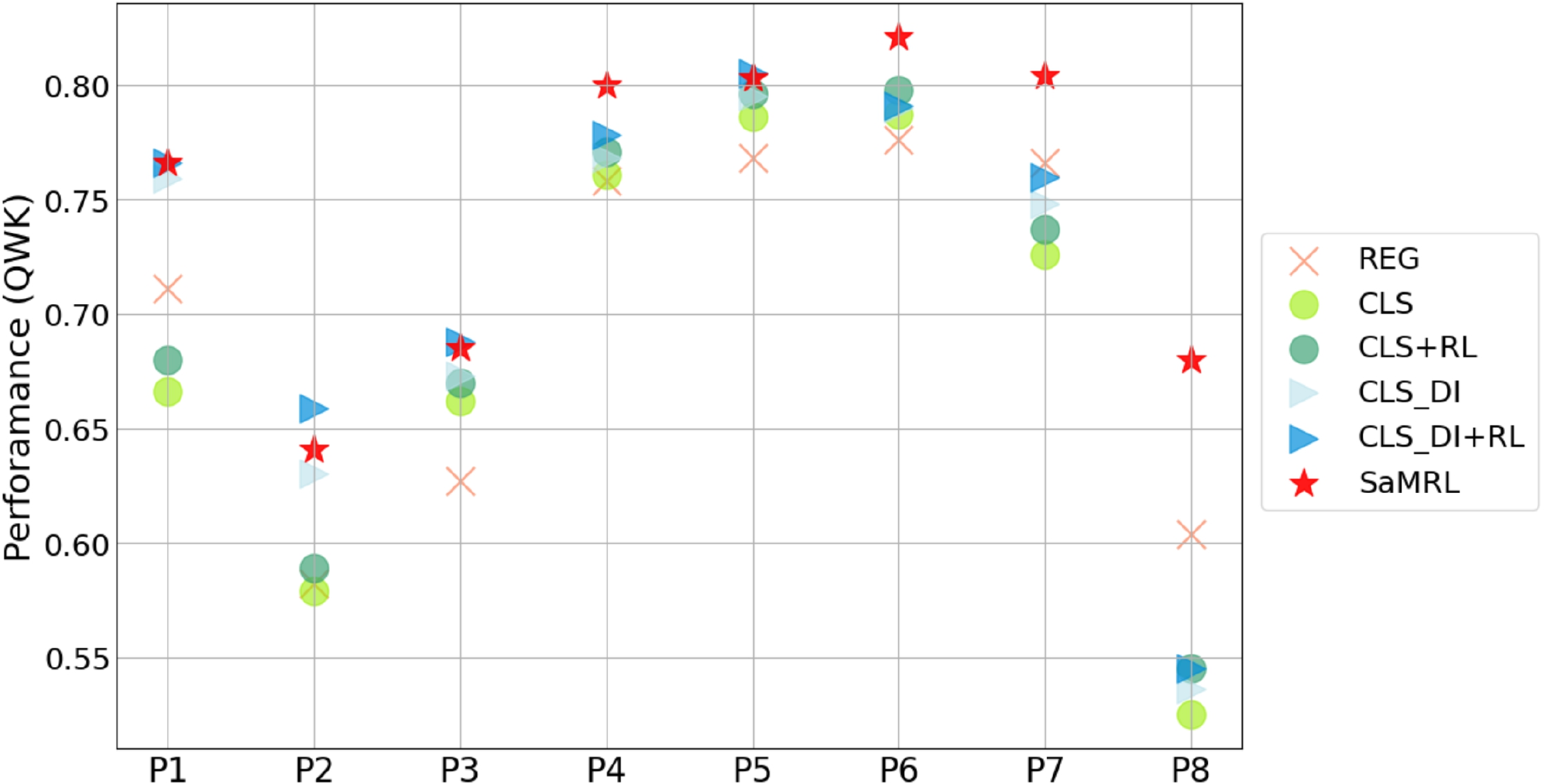}
\caption{Comparison results of classification-based RL models and our SaMRL ({\color{red}\ding{72}}) for the \textit{Overall} score prediction. $CLS_{+RL}$ and $CLS_{DI+RL}$ are models where RL is applied to $CLS$ and $CLS_{DI}$, respectively.}
\label{fig: classification}
\end{figure}

\subsection{Discussions}
\paragraph{Comparison with classification-based RL}
Our method has shown effectiveness in essay sets evaluated in a wider score range, indicating the overcoming of limitations that exist in classification-based RL methods. To investigate our actual impacts compared to them, we compare SaMRL results with existing RL-based AES systems \cite{wang-etal-2018-automatic}. Figure~\ref{fig: classification} illustrates two prior RL models: 1) bidirectional LSTM-based classification model, CLS+RL, and 2) dilated LSTM-based classification model, CLS$_{DI}$+RL. As they are holistic scoring models that predict a single \textit{Overall} score, the comparison is constrained to the \textit{Overall} trait. Our SaMRL approach outperforms the prior models in most prompts, particularly showing significant improvements in prompts 7 and 8, which have a broader rating range of [0,30] and [0,60] than other prompts. Note that in these two prompts, even when RL is applied to classification, its performance is significantly inferior to regression ({\color{orange}$\mathbf{\times}$}). Meanwhile, our model, leveraging text generation probability while incorporating an awareness of the scoring schema, demonstrates significant robustness.



\paragraph{Impact of weight learning for multi-loss}

\begin{figure}[t]
\centering
\includegraphics[width=0.97\linewidth]{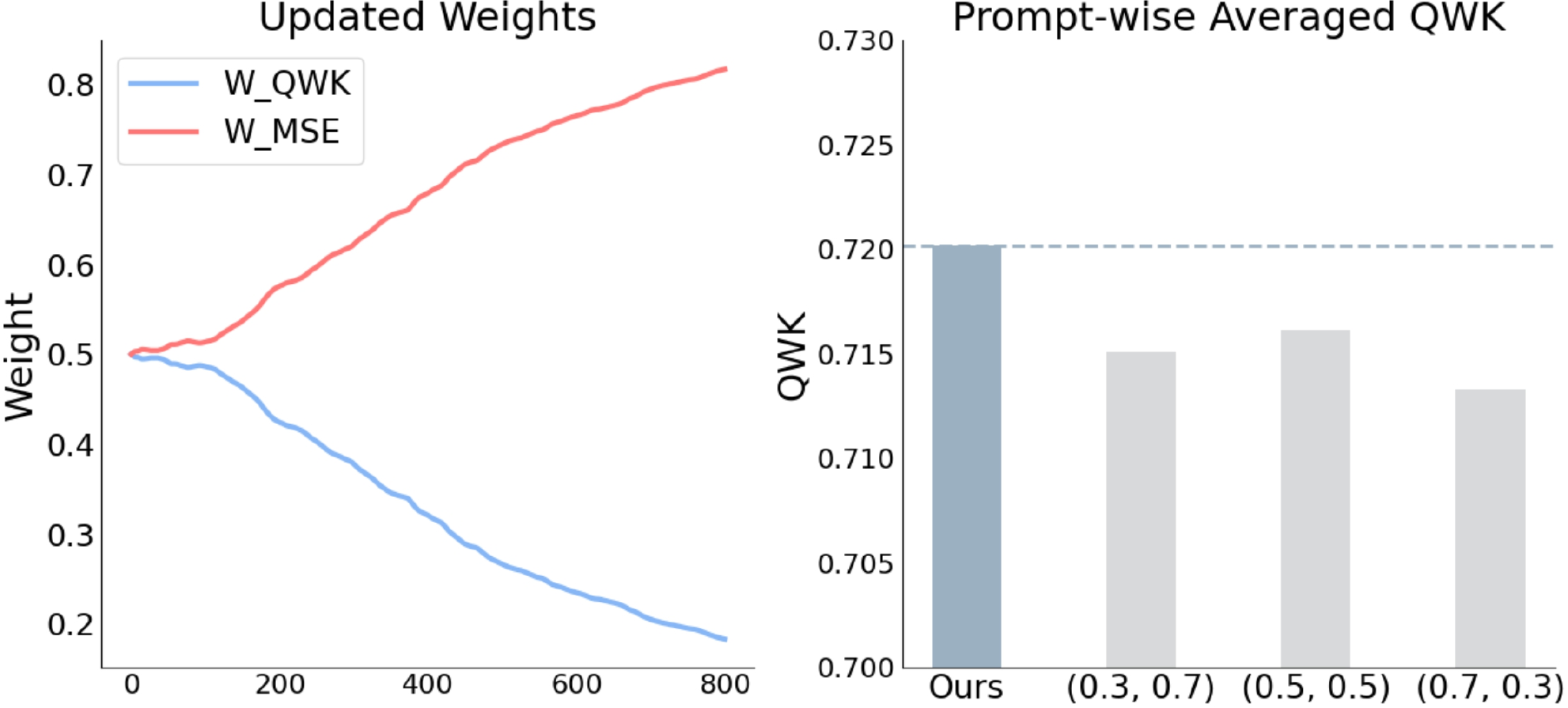}
\caption{Variations in the updated weights of loss${R_Q}$ ({\color{cyan}$W_{QWK}$}) and loss${R_M}$ ({\color{red}$W_{MSE}$}) across training steps (left); comparison of prompt-wise averaged QWK performance between models with fixed weights and our SaMRL with trainable weights.}
\label{fig: weight}
\end{figure}

Following previous research showing that adjusting the weights of each loss adaptively in multi-task learning can improve performance \cite{pmlr-v80-chen18a, kendall2018multi, mao-etal-2022-metaweighting}, we optimize multiple losses by dynamically learning the weights of each loss during the RL policy update process. By treating these weights as trainable parameters, we adaptively adjust the importance of each objective. As depicted in the left section of Figure~\ref{fig: weight}, the assigned weights for each loss dynamically adjust throughout the training steps, with a growing emphasis on the MSE loss over the QWK loss as learning progresses. This shifting trend is consistent with findings from ablation studies, which indicate a greater impact of MSE reward on assisting the training of the AES model. Furthermore, our weight update mechanism has proved superior effectiveness compared to using fixed weights of (0.3, 0.7), (0.5, 0.5), and (0.7, 0.3) for loss$_{R_Q}$ and loss$_{R_M}$, respectively, as reported in the right part of Figure~\ref{fig: weight}. We anticipate that further optimization in multi-loss weight could bring in additional performance advances in future works.


\paragraph{}

\section{Conclusion}
In this work, we propose a Scoring-aware Multi-reward Reinforcement Learning (SaMRL) method, which incorporates the rating process for effective multi-trait scoring within the generation framework. By introducing the policy gradient reinforcement in the autoregressive score generation paradigm, we enable the direct use of the QWK metric; thus, SaMRL effectively captures the rating procedure. In addition, we jointly introduce the score-level difference-based reward, MSE reward, which was previously limited to regression AES models, bringing in the overall enhancements on trait-wise and prompt-wise scoring qualities. Extensive experiments and analysis further support the assistance of our RL strategy over the simple generation. 

\section*{Limitation}
In this work, we grounded the method on autoregressive prediction, where prediction order may matter. Currently, we employ the same order as the previous work \cite{do-etal-2024-autoregressive}; however, thoroughly considering the shifts in trait prediction order can lead to further improvements. Secondly, we update the policy at a time after the entire trait score prediction, motivated by existing approaches \cite{stiennon2022learning, dutta2024frugal, ryu2024multi}. However, training the policy with the instant updating per each action (i.e., token generation) might bring in more benefits in the case of the scoring task, which can be noteworthy for future work. In addition, we update the weights via training for the multiple losses. As in prior works, adaptive optimization strategies or more refined mechanisms could extend the impact of our method \cite{pmlr-v80-chen18a, kendall2018multi, mao-etal-2022-metaweighting}. 

\section*{Ethical Statement}
Only publicly available datasets, ASAP, ASAP++, and Feedback Prize, are used in this work. 

\section*{Acknowledgements}
This work was partly supported by Institute of Information \& communications Technology Planning \& Evaluation (IITP) grant funded by the Korea government (MSIT) (No.RS-2019-II191906, Artificial Intelligence Graduate School Program (POSTECH)), the MSIT (Ministry of Science and ICT) Korea, under the ITRC (Information Technology Research Center) support program (IITP-2024-2020-0-01789) supervised by the IITP, and Institute of Information \& communications Technology Planning \& Evaluation (IITP) grant funded by the Korea government (MSIT) (No.2022-0-00223, Development of digital therapeutics to improve communication ability of autism spectrum disorder patients).

\bibliography{main}

\appendix

\begin{table*}[t]
\centering
\scalebox{
0.67}{
\begin{tabular}{l|ccccccccccc|c}
\toprule
& \multicolumn{11}{c|}{\textbf{Traits (Prediction Order: ←)}} & \\
\hline
\textbf{Model} &  Overall & Content & PA & Lang & Nar & Org & Conv & WC & SF & Style & Voice & AVG\\
\midrule
{ArTS-base*} &±0.025&	±0.015&	±0.020&	±0.027&	±0.020	&±0.021&	±0.022	&±0.033&	±0.026	&±0.008	&±0.098	& ±0.029 \\
\textbf{SaMRL-base} (Ours) & ±0.014 & ±0.015 & ±0.019 & ±0.024 & ±0.018 & ±0.019 & ±0.017 & ±0.027 & ±0.021 & ±0.007 & ±0.114 & ±0.027 \\
\midrule
{ArTS-large*} & ±0.011 & ±0.014 & ±0.032 & ±0.026 & ±0.015 & ±0.011 & ±0.016 & ±0.005 & ±0.009 & ±0.033 & ±0.080 & ±0.023 \\
\textbf{SaMRL-large} (Ours) & ±0.010 & ±0.013 & ±0.030 & ±0.023 & ±0.017 & ±0.013 & ±0.019 & ±0.009 & ±0.008 & ±0.023 & ±0.081 & ±0.022 \\
\bottomrule
\end{tabular}}
\caption{\label{table7}
The standard deviation of QWK results over five runs (main results in Table~\ref{tab: main}) averaged across the prompts for each \textbf{trait}. The \textit{AVG} here denotes the mean of the trait-specific standard deviations, while \textit{SD} in Table~\ref{tab: main} refers to the standard deviation of the five-fold average.}

\smallskip
\smallskip

\scalebox{
0.7}{
\begin{tabular}{l|cccccccc|c}
\toprule
& \multicolumn{8}{c|}{\textbf{Prompts}} & \\
\hline
\textbf{Model} & 1 & 2 & 3 & 4 & 5 & 6 & 7 & 8 & AVG \\
\hline
{ArTS-base*} & ±0.010 & ±0.034 &  ±0.044 &  ±0.020 &  ±0.018 &  ±0.011 &  ±0.007 &  ±0.089 &  ±0.029  \\
\textbf{SaMRL-base} (Ours) &  ±0.015  &  ±0.023  &  ±0.041  &  ±0.019  &  ±0.021  &  ±0.012  &  ±0.012  &  ±0.095  &  ±0.030  \\
\midrule
{ArTS-large*} &  ±0.016  &  ±0.031  &  ±0.037  &  ±0.024  &  ±0.029  &  ±0.025  &  ±0.025  &  ±0.050  &  ±0.030  \\
\textbf{SaMRL-large} (Ours) &  ±0.020  &  ±0.029  &  ±0.037  &  ±0.025  &  ±0.030  &  ±0.021  &  ±0.022  &  ±0.057  &  ±0.030  \\
\bottomrule
\end{tabular}}
\caption{\label{table8}
The standard deviation of QWK results over five runs (Table~\ref{tab: prompt}) averaged across the traits for each \textbf{prompt}.}
\end{table*}

\begin{table*}[t]
\centering
\scalebox{
0.67}{
\begin{tabular}{l|ccccccccccc|c}
\toprule
& \multicolumn{11}{c|}{\textbf{Traits (Prediction Order: ←)}} & \\
\hline
\textbf{Model} &  Overall & Content & PA & Lang & Nar & Org & Conv & WC & SF & Style & Voice & AVG\\
\hline
{ArTS-base*} &  ±0.025 & 	±0.015 & 	±0.020 	& ±0.027 & 	±0.020 	& ±0.021 & 	±0.022 	& ±0.033 & 	±0.026 	& ±0.008 	& ±0.098 	& ±0.029  \\
\midrule
{SaSRL$_M$} &  ±0.008  &  ±0.015  &  ±0.021  &  ±0.026  &  ±0.021  &  ±0.020  &  ±0.017  &  ±0.034  &  ±0.021  &  ±0.010  &  ±0.098  &  ±0.026 \\
{SaSRL$_Q$} &  ±0.024 &  ±0.016 &  ±0.022 &  ±0.029 &  ±0.028 &  ±0.019 &  ±0.016 &  ±0.030 &  ±0.024 &  ±0.015 &  ±0.077 &  ±0.027  \\
{SaMRL\_uniQ$_T$} &  ±0.027  &  ±0.015  &  ±0.017  &  ±0.025  &  ±0.020  &  ±0.022  &  ±0.020  &  ±0.032  &  ±0.026  &  ±0.010  &  ±0.098  &  ±0.028  \\
{SaMRL\_uniQ$_B$} & ±0.026& ±0.014& ±0.022& ±0.030& ±0.027& ±0.016& ±0.016 & ±0.032 & ±0.020 & ±0.010 & ±0.109 & ±0.029  \\
\midrule
\textbf{SaMRL\_Q} (Ours) & ±0.014 & ±0.015 & ±0.019 & ±0.024 & ±0.018 & ±0.019 & ±0.017 & ±0.027 & ±0.021 & ±0.007 & ±0.114 & ±0.027 \\

\bottomrule
\end{tabular}}
\caption{\label{table9}
The standard deviation of QWK results over five runs (ablation study results in Table~\ref{tab: ablation_trait}) averaged across the prompts for each \textbf{trait}. The \textit{AVG} here denotes the mean of the trait-specific standard deviations, while \textit{SD} in Table~\ref{tab: ablation_trait} refers to the standard deviation of the five-fold average.
}

\smallskip
\smallskip

\scalebox{
0.7}{
\begin{tabular}{l|cccccccc|c}
\toprule
& \multicolumn{8}{c|}{\textbf{Prompts}} & \\
\hline
\textbf{Model} & 1 & 2 & 3 & 4 & 5 & 6 & 7 & 8 & AVG \\
\hline
{ArTS-base*} &  ±0.010 &  ±0.034 &  ±0.044 &  ±0.020 &  ±0.018 &  ±0.011 &  ±0.007 &  ±0.089 &  ±0.029  \\
\midrule
{SaSRL$_M$} & ±0.018 & ±0.027 & ±0.044 & ±0.019 & ±0.016 & ±0.012 & ±0.012 & ±0.087 & ±0.029 \\
{SaSRL$_Q$} &  ±0.017  &  ±0.032  &  ±0.050  &  ±0.022  &  ±0.013  &  ±0.016  &  ±0.012  &  ±0.078  &  ±0.030  \\
{SaMRL\_uniQ$_T$} &  ±0.010  &  ±0.033  &  ±0.040  &  ±0.022  &  ±0.020  &  ±0.010  &  ±0.008  &  ±0.090  &  ±0.029  \\
{SaMRL\_uniQ$_B$} &  ±0.019  &  ±0.035  &  ±0.049  &  ±0.021  &  ±0.021  &  ±0.014  &  ±0.012  &  ±0.089  &  ±0.033  \\

\midrule
\textbf{SaMRL\_Q} (Ours) &  ±0.015  &  ±0.023  &  ±0.041  &  ±0.019  &  ±0.021  &  ±0.012  &  ±0.012  &  ±0.095  &  ±0.030 \\

\bottomrule
\end{tabular}}
\caption{\label{table10}
The standard deviation of QWK results over five runs (Table~\ref{tab: ablation_prompt}) averaged across the traits for each \textbf{prompt}.}
\end{table*}

\section{Comparison models}
\label{Comparison models}
We primarily compared our method with the baseline ArTS \cite{do-etal-2024-autoregressive} model, which is the previous state-of-the-art model for multi-trait AES. As we aim to examine the effects of applying RL on the autoregressive model, the comparison mainly focuses on the model with and without applying our method. In addition, we also report the results of other multi-trait scoring models \cite{kumar2022many} and the holistic scoring models \cite{cozma-etal-2018-automated, dong2017attention} individually applied for each trait prediction. In particular, the multi-trait scoring MTL model \cite{kumar2022many} constructed each trait-specific layer and used all other trait layers auxiliary for a target trait training and prediction. The holistic scoring model, HISK, utilizes a support vector regressor paired with a histogram intersection string kernel, whereas STL-{\small LSTM} models use an LSTM-CNN-based structure; each model is iteratively deployed for independent trait scoring task. Except for the implemented ArTS*, results of other models are reported from the previous works \cite{kumar2022many, do-etal-2024-autoregressive}.

\section{Standard deviations for five-fold validations}
\label{sec:appendix}
In this work, we implemented five-fold validation and presented scores averaged over the five folds in all experiments. In this session, we report the standard deviation for the main results (Table~\ref{table7}, \ref{table8}) and the results of ablation studies (Table~\ref{table9}, \ref{table10}).


\end{document}